\documentclass[11pt]{article}
\usepackage{acl2015}
\usepackage{url}
\usepackage{times}
\usepackage{latexsym}
\usepackage{graphicx}
\usepackage{algorithm2e}
\usepackage{booktabs}
\usepackage{paralist}
\usepackage{color}
\usepackage{pdfsync}
\usepackage{enumerate}
\usepackage{mathtools}
\usepackage{tikz}
\usepackage{tikz}
\usetikzlibrary{positioning,shapes,snakes,backgrounds}
\usepackage[scale=0.82, top=1.5cm, bottom=2.5cm]{geometry}
\usepackage[absolute]{textpos}
\newcommand{\email}[1]{{\footnotesize\url{#1}}}

\newcommand{\ra}[1]{\renewcommand{\arraystretch}{#1}}
\usepackage{xspace,multirow}

\newcommand{\np}{---}
\newcommand{\nr}{???}
\newcommand{\intr}{$\times\!\times\!\times$}

\newcommand{\kd}{$k$-d\xspace}

\newcommand{\tabref}[2][]{Table#1~\ref{#2}\xspace}
\newcommand{\figref}[2][]{Figure#1~\ref{#2}\xspace}

\newcommand{\dataset}[1]{\textsc{#1}\xspace}
\newcommand{\GEOTEXT}{\dataset{GeoText}}
\newcommand{\TwitterUS}{\dataset{Twitter-US}}
\newcommand{\TwitterWorld}{\dataset{Twitter-World}}

\newcommand{\method}[1]{\texttt{#1}\xspace}
\newcommand{\LR}{\method{LR}}
\newcommand{\LP}{\method{LP}}
\newcommand{\LPLR}{\method{LP-LR}}
\newcommand{\MAD}{\method{MAD-B}}
\newcommand{\MADCELB}{\method{MADCEL-B}}
\newcommand{\MADCELW}{\method{MADCEL-W}}
\newcommand{\MADCELBLR}{\method{MADCEL-B-LR}}
\newcommand{\MADCELWLR}{\method{MADCEL-W-LR}}

\newcommand{\class}[1]{\texttt{#1}\xspace}
\newcommand{\unknown}{\class{Unknown}}

\newcommand{\nearacc}{\textsf{Acc@161}\xspace}
\newcommand{\Mean}{\textsf{Mean}\xspace}
\newcommand{\Median}{\textsf{Median}\xspace}

\date{}

\title{Twitter User Geolocation Using a Unified Text and Network Prediction Model}

\author{Afshin Rahimi, Trevor Cohn, \and Timothy Baldwin\\
  Department of Computing and Information Systems\\
        The University of Melbourne\\
  \email{arahimi@student.unimelb.edu.au}\\
  \email{{t.cohn,tbaldwin}@unimelb.edu.au}}

\date{}

\begin{document}

\maketitle

\begin{abstract}
  We propose a label propagation approach to geolocation prediction
  based on Modified Adsorption, with two enhancements: (1) the removal
  of ``celebrity'' nodes to increase location homophily and boost
  tractability; and (2) the incorporation of text-based geolocation
  priors for test users. Experiments over three Twitter benchmark
  datasets achieve state-of-the-art results, and demonstrate the
  effectiveness of the enhancements.
\end{abstract}

\begin{textblock}{0.8}[0.5,0.5](0.5,0.96) \begin{center} \noindent
\small
\textit{
Proceedings of the 53rd Annual Meeting of the Association for Computational Linguistics and 
the 7th International Joint Conference on Natural Language Processing (Short Papers), 
pages 630-636, Beijing, China, July 26-31, 2015. c 2015 Association for Computational Linguistics}
\end{center}
\end{textblock}

\section{Introduction}
Geolocation of social media users is essential in applications ranging
from rapid disaster response \cite{earle2010omg,ashktorab2014tweedr,morstatter2013understanding} and
opinion analysis \cite{mostafa2013more,kirilenko2014public}, to
recommender systems \cite{noulas2012random,schedl2014location}.  Social
media platforms like Twitter provide support for users to declare their
location manually in their text profile or automatically with GPS-based
geotagging. However, the text-based profile locations are noisy and only
1--3\% of tweets are geotagged \cite{cheng2010you,morstatter2013sample},
meaning that geolocation needs to be inferred from other information
sources such as the tweet text and network relationships.

User geolocation is the task of inferring the primary (or ``home'') location
of a user from available sources of information, such as text posted by
that individual, or network relationships with other
individuals \cite{han2014text}. Geolocation models are usually trained on the small set of
users whose location is known (e.g.\ through GPS-based geotagging), and
other users are geolocated using the resulting model.  These models
broadly fall into two categories: text-based and network-based
methods. Orthogonally, the geolocation task can be viewed as a
regression task over real-valued geographical coordinates, or a
classification task over discretised region-based locations.

Most previous research on user geolocation has focused either on
text-based classification approaches
\cite{eisenstein2010latent,wing2011simple,roller2012supervised,han2014text}
or, to a lesser extent, network-based regression
approaches~\cite{jurgens2013s,compton2014geotagging,rahimi2015exploiting}. Methods
which combine the two, however, are rare.

In this paper, we present our work on Twitter user geolocation using both
text and network information. Our contributions are as follows:
(1) we propose the use of Modified Adsorption~\cite{talukdar2009} as a baseline
network-based geolocation model, and show that it outperforms previous network-based
approaches \cite{jurgens2013s,rahimi2015exploiting};
(2) we demonstrate that removing ``celebrity'' nodes (nodes with high
in-degrees) from the network increases geolocation accuracy and dramatically
decreases network edge size; and
(3) we integrate text-based geolocation priors into Modified Adsorption,
and show that our unified geolocation model
outperforms both text-only and network-only approaches, and achieves
state-of-the-art results over three standard datasets.

\section{Related Work}
\label{sec:related}

A recent spike in interest on user geolocation over social media data
has resulted in the development of a range of approaches to automatic geolocation prediction, based on information
sources such as the text of messages, social networks, user profile data,
and temporal data. 
Text-based methods model the geographical bias of language use in social
media, and use it to geolocate non-geotagged users.
Gazetted expressions~\cite{leidner2011detecting} and geographical names~\cite{quercini2010determining}
were used as feature in early work, but were shown to be sparse in coverage.
\newcite{han2014text} used information-theoretic methods to
automatically extract location-indicative words for location
classification. 
\newcite{wing2014hierarchical} reported that discriminative approaches
(based on hierarchical classification over adaptive grids), when
optimised properly, are superior to explicit feature selection.
\newcite{cha2015twitter} showed that sparse coding can be used to effectively learn a latent
representation of tweet text to use in user geolocation.
\newcite{eisenstein2010latent} and \newcite{ahmed2013hierarchical}
proposed topic model-based approaches to geolocation, based on the
assumption that words are generated from hidden topics and geographical
regions.
Similarly, \newcite{yuan2013and} used graphical models to jointly learn
spatio-temporal topics for users. The advantage of these generative
approaches is that they are able to work with the continuous
geographical space directly without any pre-discretisation, but
they are algorithmically complex and don't scale well to larger
datasets. \newcite{hulden2015kernel} used kernel-based methods to
smooth linguistic features over very small grid sizes to alleviate
data sparseness.

Network-based geolocation models, on the other hand, utilise the fact
that social media users interact more with people who live
nearby.
\newcite{jurgens2013s} and \newcite{compton2014geotagging} used a
Twitter reciprocal mention network, and geolocated users based on the
geographical coordinates of their friends, by minimising the weighted
distance of a given user to their friends. 
For a reciprocal mention network to be effective, however, a huge amount of Twitter
data is required. 
\newcite{rahimi2015exploiting} showed that this
assumption could be relaxed to use an undirected mention network for
smaller datasets, and still attain state-of-the-art results.
The greatest shortcoming of network-based models is that
they completely fail to geolocate users who are not connected to
geolocated components of the graph. 
As shown by \newcite{rahimi2015exploiting}, geolocation predictions from
text can be used as a backoff for disconnected users, but there has been little work that has
investigated a more integrated text- and network-based approach to user
geolocation.

\section{Data}
\label{sec:data}

We evaluate our models over three pre-existing geotagged Twitter
datasets: (1) \GEOTEXT~\cite{eisenstein2010latent}, (2) \TwitterUS~\cite{roller2012supervised},
and (3) \TwitterWorld~\cite{bo2012geolocation}.
In each dataset, users are represented by a single meta-document,
generated by concatenating their tweets.
The datasets are pre-partitioned into training, development and test sets, and
rebuilt from the original version to include mention information. The
first two datasets were constructed to contain mostly English messages.

\GEOTEXT consists of tweets from 9.5K users: 1895 users are held out for
each of development and test data. The primary location of each user is set to
the coordinates of their first tweet. 

\TwitterUS consists of 449K users, of which 10K users are held out for
each of development and test data. The primary location of each user is, once
again, set to the coordinates of their first tweet.

\TwitterWorld consists of 1.3M users, of which 10000 each are held out
for development and test.  Unlike the other two datasets, the primary
location of users is mapped to the geographic centre of the city where the majority
of their tweets were posted.

\section{Methods}
\label{sec:methods}

We use label propagation over an @-mention graph in our models.
We use \kd tree descretised adaptive grids as class labels for users
and learn a label distribution for each user by label propagation
over the @-mention network using labelled nodes as seeds. For \kd tree discretisation, we set the number
of users in each region to 50, 2400, 2400 for \GEOTEXT, \TwitterUS and \TwitterWorld
respectively, based on tuning over the development data.

\begin{figure*}[t]
  \centering
     \includegraphics[width=\textwidth]{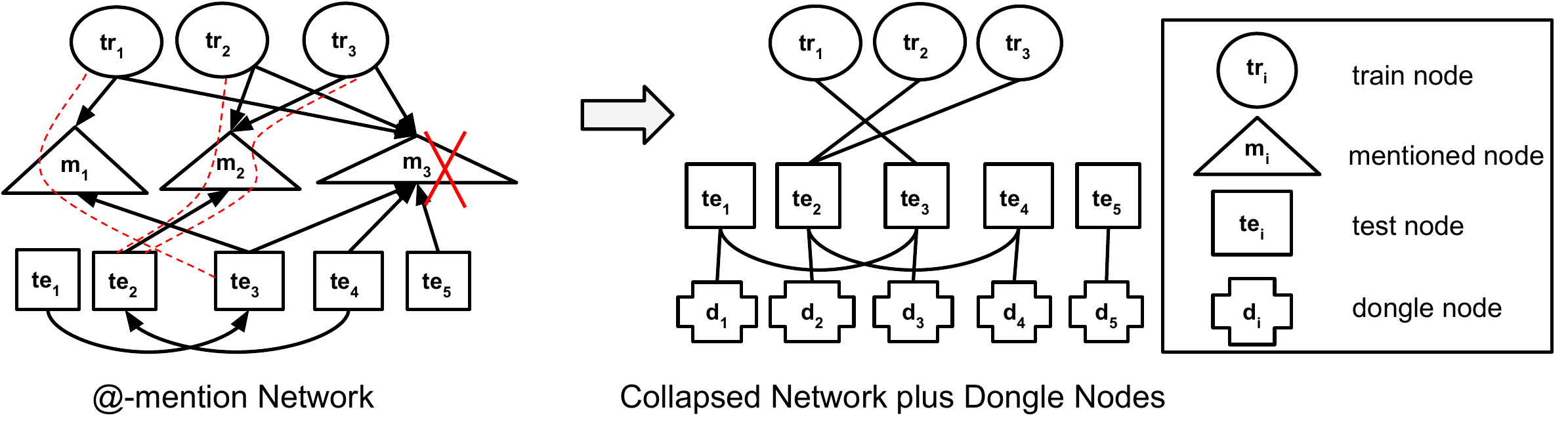}
     \caption{A collapsed network is built from the @-mention
       network. Each mention is shown by a directed arrow, noting that
       as it is based exclusively on the tweets from the training and
       test users, it will always be directed from a training or test
       user to a mentioned node.  All mentioned nodes which are not a
       member of either training or test users are removed and the
       corresponding training and test users, previously connected
       through that node, are connected directly by an edge, as
       indicated by the dashed lines.  Mentioned nodes with more than
       $T$ unique mentions (celebrities, such as $m_3$) are removed from
       the graph.  To each test node, a dongle node that carries the
       label from another learner (here, text-based \LR) is added in
       \MADCELBLR and \MADCELWLR.}
  \label{fig:network}
\end{figure*}

\paragraph*{Social Network:}
We used the @-mention information to build an undirected graph
between users. In order to make the inference more tractable, we
removed all nodes that were not a member of the training/test set, and
connected all pairings of training/test users if there was any path
between them (including paths through non training/test users). We
call this network a ``collapsed network'', as illustrated in
\figref{fig:network}. Note that a celebrity node with $n$ mentions connects $n(n-1)$ nodes in the collapsed
network. We experiment with both binary
and weighted edge (based on the number of mentions connecting the given users) networks.

\paragraph*{Baseline:}
Our baseline geolocation model (``\MAD'') is formulated as label
propagation over a binary collapsed network, based on Modified
Adsorption~\cite{talukdar2009}.
It applies to a graph $G=(V, E, W)$ where $V$ is the set of nodes with
$|V|=n=n_l+n_u$  (where $n_l$ nodes are labelled and $n_u$ nodes
are unlabelled), $E$ is the set of edges, and $W$ is an edge weight
matrix. Assume $C$ is the set of labels where $|C|=m$ is
the total number of labels. $Y$ is an $n\times m$ matrix storing
the training node labels, and $\hat{Y}$ is the estimated label distribution
for the nodes. The goal is to estimate $\hat{Y}$
for all nodes (including training nodes) so that the following objective
function is minimised:
\begin{equation*}
 \begin{aligned}
 \displaystyle C(\hat{Y}) = \sum_{l} \bigg{[} \,  \mu_{1} (Y_l - \hat{Y}_l)^T S (Y_l - \hat{Y}_l) + \\ 
 \mu_{2} \hat{Y}_l^T L \hat{Y}_l\bigg{]} 
 \end{aligned}
\end{equation*}
where $\mu_1$ and $\mu_2$ are hyperparameters;\footnote{In the base
  formulation of \MAD, there is also a regularisation term with weight
  $\mu_3$, but in all our experiments, we found that the best results
  were achieved over development data with $\mu_3 = 0$, i.e.\ with no regularisation; the term is
  thus omitted from our description.} $L$
is the Laplacian of an undirected graph derived from $G$; and $S$
is a diagonal binary matrix indicating if a node is labelled
or not. 
The first term of the equation forces the labelled nodes to keep their
label (prior term), while the second term pulls a node's label toward that of
its neighbours (smoothness term). 
For the first term,
the label confidence for training and test users is set to 1.0 and 0.0,
respectively.
Based on the development data, we set $\mu_{1}$ and $\mu_{2}$ 
to 1.0 and 0.1, respectively, for all the experiments. 
For \TwitterUS and \TwitterWorld, the inference was intractable for the default network, as it
was too large.\\

There are two immediate issues with the baseline graph propagation
method: (1) it doesn't scale to large datasets with 
high edge counts, related to which, it tends to be biased by highly-connected nodes; and (2) it can't predict the geolocation of test users who aren't connected to any training
user (\MAD returns \unknown, which we rewrite with the centre of the map). We redress these two issues as follows.

\begin{table*}[t!]
\centering
\ra{1.1}
\smallskip\noindent
\resizebox{\linewidth}{!}{%
\begin{tabular}{@{}lccccccccccc@{}}\toprule
& \multicolumn{3}{c}{\GEOTEXT} & \phantom{abc}& \multicolumn{3}{c}{\TwitterUS} &
\phantom{abc} & \multicolumn{3}{c}{\TwitterWorld}\\
\cmidrule{2-4} \cmidrule{6-8} \cmidrule{10-12}
                                                   & \nearacc & \Mean      & \Median    &&  \nearacc & \Mean    & \Median      &&  \nearacc & \Mean     & \Median  \\ 
\MAD                                               & 50       & 683        & 146        && \intr     & \intr    & \intr        &&  \intr    & \intr     & \intr    \\
\MADCELB                                           & 56       & 609        & 76         && 54        & 709      & 117          &&  70       & 936       & \bf{0}   \\
\MADCELW                                           & 58       & 586        & 60         && 54        & 705      & 116          &&  71       & 976       & \bf{0}   \\
\MADCELBLR                                         & 57       & 608        & 65         && \bf{60}   & 533      & \bf{77}      &&  \bf{72}  & \bf{786}  & \bf{0}   \\
\MADCELWLR                                         & \bf{59}  & \bf{581}   & \bf{57}    && \bf{60}   & \bf{529} & 78           &&  \bf{72}  & 802       & \bf{0}   \\
\midrule
\LR~\cite{rahimi2015exploiting}                    & 38       & 880        & 397        &&  50       & 686      & 159          &&  63       & 866       & 19       \\
\LP~\cite{rahimi2015exploiting}                    & 45       & 676        & 255        &&  37       & 747      & 431          &&  56       & 1026      & 79       \\
\LPLR~\cite{rahimi2015exploiting}                  & 50       & 653        & 151        &&  50       & 620      & 157          &&  59       & 903       & 53       \\
\newcite{wing2014hierarchical} (uniform)           & \np      & \np        & \np        &&  49       & 703      & 170          &&  32       & 1714      & 490      \\
\newcite{wing2014hierarchical} (\kd)               & \np      & \np        & \np        &&  48       & 686      & 191          &&  31       & 1669      & 509      \\
\newcite{bo2012geolocation}                        & \np      & \np        & \np        && 45        & 814      & 260          &&  24       & 1953      & 646      \\ 
\newcite{ahmed2013hierarchical}                    & \nr      & \nr        & 298        && \np       & \np      & \np          &&  \np      & \np       & \np      \\
\newcite{cha2015twitter}                           & \nr      & \bf{581}   & 425        && \np       & \np      & \np          &&  \np      & \np       & \np      \\
\bottomrule
\end{tabular}
}
\caption{Geolocation results over the 
  three Twitter corpora, comparing baseline Modified Adsorption (\MAD),
  with Modified Adsorption with celebrity removal (\MADCELB and \MADCELW, over binary
  and weighted networks, resp.) or celebrity removal plus text priors
  (\MADCELBLR and \MADCELWLR, over binary and weighted networks, resp.);
  the table also includes state-of-the-art results for each dataset (``\np''
  signifies that no results were published for the given dataset;
  ``\nr'' signifies that no results were reported for the given metric;
  and ``\intr'' signifies that results could not be generated, due to
  the intractability of the training data).
}
\label{tab:results}
\end{table*}

\begin{figure}[t]
  \centering
  \includegraphics[scale=1,width=\linewidth]{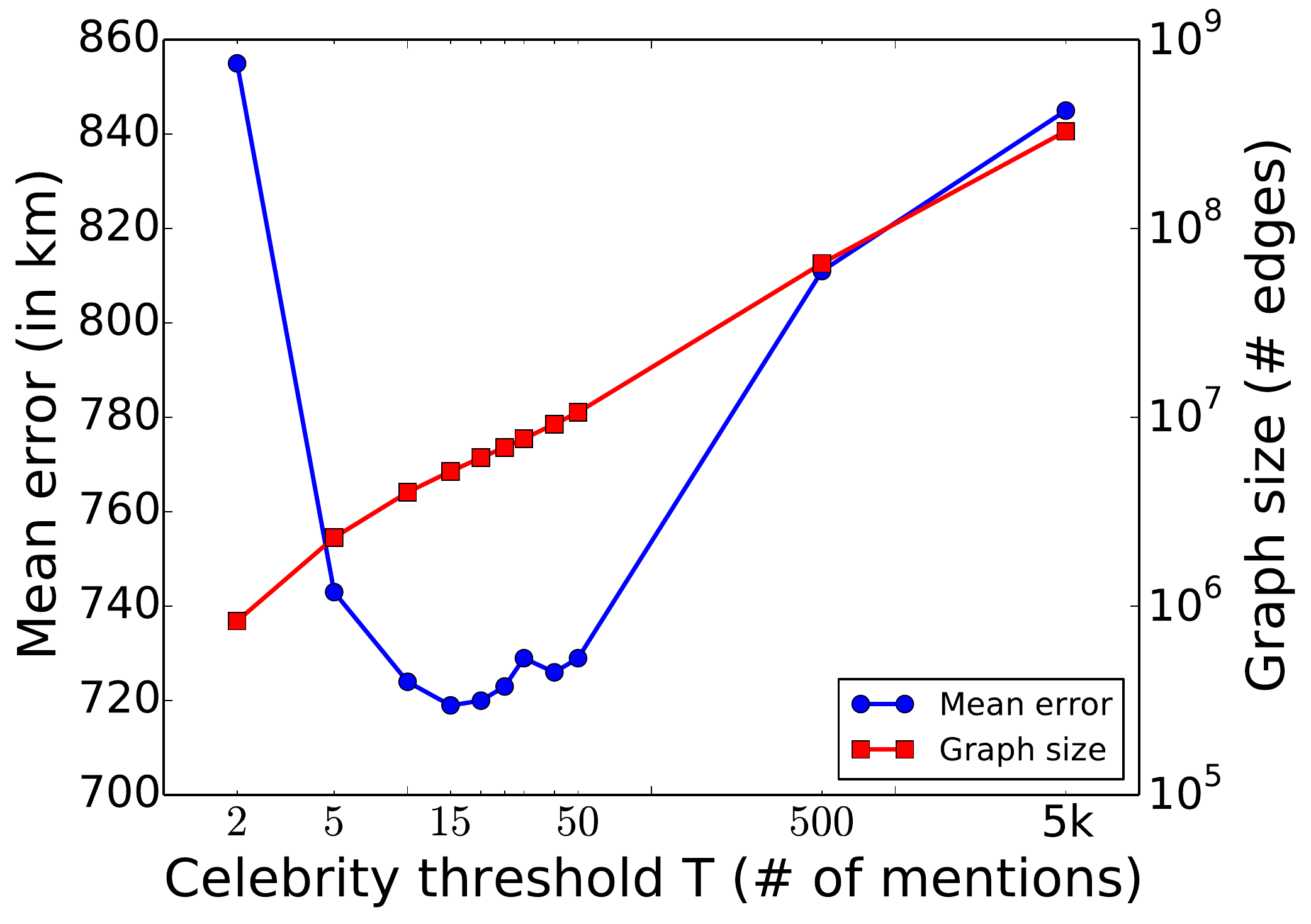}
  \caption{Effect of celebrity removal on geolocation performance and graph size. For each $T$ performance is
  measured over the development set of \TwitterUS by \MADCELW.}
  \label{fig:celebrity}
\end{figure}

\paragraph{Celebrity Removal}
To address the first issue, we target ``celebrity'' users, i.e.\ highly-mentioned
Twitter users. 
Edges involving these users often carry little or no geolocation information
(e.g.\ the majority of people who mention Barack Obama don't live in
Washington D.C.).
Additionally, these users tend to be highly connected to other users and
generate a disproportionately high number of edges in the graph, leading
in large part to the baseline \MAD not scaling over large datasets such as
\TwitterUS and \TwitterWorld.
We identify and filter out celebrity nodes simply by assuming that a celebrity
is mentioned by more than $T$ users, where $T$ is tuned over development
data.
Based on tuning over the development set of \GEOTEXT and \TwitterUS, $T$ was set to 5 and 15 respectively.
For \TwitterWorld tuning was very resource intensive so $T$ was set to 5
based on \GEOTEXT, to make the inference faster.
Celebrity removal dramatically reduced the edge count in all three datasets (from $1 \times 10^9$ to $5 \times 10^6$ for \TwitterUS and from $4
\times 10^{10}$ to $1 \times 10^7$ for \TwitterWorld), and made inference tractable 
for \TwitterUS and \TwitterWorld.~\newcite{jurgens2015geolocation}
report that the time complexity of most network-based geolocation methods is $\mathcal{O}(k^2)$ for each node where $k$ is the average 
number of vertex neighbours. In the case of the collapsed network of
\TwitterWorld, $k$ is decreased by a factor of $4000$ after
setting the celebrity threshold $T$ to $5$.
We apply celebrity removal over both binary
(``\MADCELB'') and weighted (``\MADCELW'') networks (using the respective $T$ for each dataset). The effect of celebrity
removal over the development set of \TwitterUS is shown in \figref{fig:celebrity} where
it dramatically reduces the graph edge size and simultaneously leads to
an improvement in the mean error.

\paragraph{A Unified Geolocation Model}
To address the issue of disconnected test users, we incorporate text
information into the model by attaching a labelled dongle node to every
test node~\cite{zhu2002learning,goldberg2006seeing}.
The label for the dongle node is based on a text-based $l_1$
regularised logistic regression model, using the method
of~\newcite{rahimi2015exploiting}. 
The dongle nodes with their corresponding label confidences are added to
the seed set, and are treated in the same way as other labelled
nodes (i.e.\ the training nodes). 
Once again, we experiment with text-based labelled dongle nodes over
both binary (``\MADCELBLR'') and weighted (``\MADCELWLR'') networks.

\section{Evaluation}
\label{sec:evaluation}

Following \newcite{cheng2010you} and \newcite{eisenstein2010latent}, we
evaluate using the mean  and median error (in km) over
all test users (``\Mean'' and ``\Median'', resp.), and also accuracy within 161km of the actual
location (``\nearacc'').
Note that higher numbers are better for \nearacc, but lower numbers are better for mean 
and median error, with a lower bound of $0$ and no
(theoretical) upper bound. 

To generate a continuous-valued latitude/longitude coordinate for a
given user from the \kd tree cell, we use the median coordinates of all
training points in the predicted region.

\section{Results}

\tabref{tab:results} shows the performance of \MAD, \MADCELB, \MADCELW, \MADCELBLR and
\MADCELWLR over the \GEOTEXT, \TwitterUS and \TwitterWorld datasets.  The results are also
compared with prior work on network-based geolocation using label propagation
(\LP)~\cite{rahimi2015exploiting}, text-based classification
models~\cite{bo2012geolocation,wing2011simple,wing2014hierarchical,rahimi2015exploiting,cha2015twitter},
text-based graphical models~\cite{ahmed2013hierarchical}, and
network--text hybrid models (\LPLR)~\cite{rahimi2015exploiting}.

Our baseline network-based model of \MAD outperforms the text-based
models and also previous network-based models~\cite{jurgens2013s,compton2014geotagging,rahimi2015exploiting}.
The inference, however, is intractable for \TwitterUS and \TwitterWorld due to the size of the network.

Celebrity removal in \MADCELB and \MADCELW has a positive effect on
geolocation accuracy, and results in a 47\% reduction
in \Median over \GEOTEXT. It also makes
graph inference over \TwitterUS and \TwitterWorld tractable, and results in superior
\nearacc and \Median, but slightly inferior \Mean, compared to the
state-of-the-art results of \LR, based on text-based classification \cite{rahimi2015exploiting}.

\MADCELW (weighted graph) outperforms \MADCELB (binary graph) over the
smaller \GEOTEXT dataset where it compensates for the sparsity of
network information, but doesn't improve the results for the two larger
datasets where network information is denser.

Adding text to the network-based geolocation models in the form of
\MADCELBLR (binary edges) and \MADCELWLR (weighted edges), we achieve
state-of-the-art results over all three datasets.  The inclusion of
text-based priors has the greatest impact on \Mean, resulting in an additional 26\% and 23\% error
reduction over \TwitterUS and \TwitterWorld, respectively. The reason
for this is that it provides a
user-specific geolocation prior for (relatively) disconnected users.

\section{Conclusions and Future Work}

We proposed a label propagation method over adaptive grids based on
collapsed @-mention networks using Modified Adsorption, and successfully
supplemented the baseline algorithm by: (a) removing ``celebrity''
nodes (improving the results and also making inference more tractable);
and (b) incorporating text-based geolocation priors into the model. 

As future work, we plan to use temporal data and also look at improving the 
text-based geolocation model using sparse
coding~\cite{cha2015twitter}. We also plan to investigate more nuanced
methods for differentiating between global and local celebrity nodes, to
be able to filter out global celebrity nodes but preserve local nodes
that can have high geolocation utility.

\subsection*{Acknowledgements}

We thank the anonymous reviewers for their insightful comments and
valuable suggestions. This work was funded in part by the Australian
Research Council.

\bibliographystyle{acl}
\bibliography{Master}

\end{document}